\documentclass[lettersize,journal]{IEEEtran}
\usepackage{amsmath,amsfonts}
\usepackage{algorithmic}
\usepackage{algorithm}
\usepackage{array}
\usepackage[caption=false,font=footnotesize]{subfig}
\usepackage{textcomp}
\usepackage{stfloats}
\usepackage{url}
\usepackage{verbatim}
\usepackage{graphicx}
\usepackage{cite}
\usepackage{orcidlink}
\usepackage{multirow}
\usepackage{array}
\usepackage{amsmath,mathtools,breqn}
\usepackage{hyperref}
\usepackage{overpic}
\begin{document}

\title{SDiFL: Stable Diffusion-Driven Framework for Image Forgery Localization}

\author{Yang Su$^{\orcidlink{0000-0002-7876-1462}}$, Shunquan Tan*$^{\orcidlink{0000-0002-7457-3691}}$, \IEEEmembership{Senior Member, IEEE}, Jiwu Huang$^{\orcidlink{0000-0002-7625-5689}}$, \IEEEmembership{Fellow, IEEE}
}

\markboth{Journal of \LaTeX\ Class Files,~Vol.~14, No.~8, August~2021}%
{Shell \MakeLowercase{\textit{et al.}}: A Sample Article Using IEEEtran.cls for IEEE Journals}


\maketitle

\begin{abstract}
Driven by the new generation of multi-modal large models, such as Stable Diffusion (SD), image manipulation technologies have advanced rapidly, posing significant challenges to image forensics. However, existing image forgery localization methods, which heavily rely on labor-intensive and costly annotated data, are struggling to keep pace with these emerging image manipulation technologies. To address these challenges, we are the first to integrate both image generation and powerful perceptual capabilities of SD into an image forensic framework, enabling more efficient and accurate forgery localization. First, we theoretically show that the multi-modal architecture of SD can be conditioned on forgery-related information, enabling the model to inherently output forgery localization results. Then, building on this foundation, we specifically leverage the multi-modal framework of Stable DiffusionV3 (SD3) to enhance forgery localization performance. We leverage the multi-modal processing capabilities of SD3 in the latent space by treating image forgery residuals—high-frequency signals extracted using specific high-pass filters—as an explicit modality. This modality is fused into the latent space during training to enhance forgery localization performance. Notably, our method fully preserves the latent features extracted by SD3, thereby retaining the rich semantic information of the input image. Experimental results show that our framework achieves up to 12\% improvements in performance on widely used benchmarking datasets compared to current state-of-the-art image forgery localization models. Encouragingly, the model demonstrates strong performance on forensic tasks involving real-world document forgery images and natural scene forging images, even when such data were entirely unseen during training.
\end{abstract}


\bstctlcite{sdifl_bstctl}

\section{Introduction}
\IEEEPARstart{T}{raditional} image forgery techniques—such as splicing, copy-move, and inpainting—have long posed a threat to the credibility of digital media \cite{mehrjardi2023survey}. These classical manipulation methods involve forging with image content by adding, removing, or duplicating regions within an image, often without introducing easily noticeable visual artifacts. As a result, they can deceive both human observers and automated detection systems \cite{singh2024image, kaur2024passive}. 

This challenge has been further exacerbated by the emergence of AI-generated image forgeries, particularly those produced by advanced models like Stable Diffusion (SD) \cite{rombach2022high}. Leveraging the latent space, SD can generate images with superior visual quality. Additionally, its multi-modal integration—particularly of text and image modalities—allows for the synthesis of highly diverse content, further challenging the reliability of visual media \cite{shen2023diffusion}.

In response to these evolving threats, considerable research efforts have been devoted to advancing image forgery localization, which has undergone continuous methodological evolution in recent years. Broadly, this evolution can be characterized by three major phases. Initially, CNN-based approaches dominated the field, introducing innovations such as bridged architectures for spatial localization \cite{zhuang2021image}, self-adversarial training strategies with coarse-to-fine networks \cite{zhuo2022self}, DCT-coefficient processing for improved compression resilience \cite{kwon2021cat}, and noise-injected U-Nets for enhanced robustness \cite{wu2022robust}. Subsequently, hierarchical frameworks were proposed to capture multi-scale forgery patterns more effectively \cite{guo2023hierarchical}.

Transformer-based methods subsequently gained prominence, exploiting self-attention mechanisms for global context modeling. Patch-based processing and dual-stage training strategies improved manipulation tracing \cite{li2023transformer}, while a dual-branch Transformer network integrating RGB and noise domain features further enhanced performance \cite{liu2023tbformer}. Hybrid architectures that combine image and noise features also boosted detection precision \cite{guillaro2023trufor}. More recently, MGQFormer introduced a mask-guided query-based Transformer framework that leverages guiding query tokens and a mask-guided loss to significantly improve both training efficiency and localization accuracy \cite{zeng2024mgqformer}.

Recent efforts have explored more advanced paradigms for forgery localization. Reinforcement learning-based approaches model the task as a pixel-level Markov Decision Process, where agents iteratively refine forgery probabilities through sequential decision-making \cite{Peng2024CoDE}. Meanwhile, diffusion models have emerged as a promising direction: Xu et al. \cite{xu2024diff} proposed a coarse-to-fine framework incorporating a pre-trained feature extractor and diffusion-based refinement, while Yu et al. \cite{Yu2024CVPR} introduced a self-supervised denoising diffusion approach that eliminates the need for manual annotations. In addition, a recent training-free method uses diffusion models to compare reconstructions, generating manipulation localization maps without supervision \cite{zhang2025training}.

While image forgery localization methods have achieved encouraging progress, existing approaches continue to face two fundamental challenges. First, the acquisition of high-quality forged image samples remains a significant bottleneck. Second, most existing methods fail to fully exploit the powerful image understanding capabilities of large pre-trained models \cite{liu2024forgerygpt}. As a result, they struggle to generalize to the diverse and rapidly evolving manipulations enabled by modern forgery methods, leading to limited robustness and generalization in real-world scenarios.

The data bottleneck in forgery localization stems from two flawed paradigms: labor-intensive manual forgery creation and unrealistic automated synthesis. This scarcity hampers progress by forcing forensic models to perpetually lag behind the evolution of forgery techniques. A paradigm shift is thus imperative. SD, with its strong capacity for modeling the underlying semantics of visual content, offers a promising alternative. By incorporating forgery-related cues into the semantic conditioning process—analogous to how text prompts guide image synthesis—SD can be repurposed to generate precise forgery localization maps. In doing so, the generation process itself is transformed from a content creation tool into a verification asset, effectively bridging the gap between advanced forgery techniques and forensic detection models.

Our proposed framework is fundamentally built upon SD and adopts the above-mentioned new paradigm. This paradigm differs fundamentally from existing forgery localization approaches~(even including those prior SD based ones~\cite{xu2024diff,Yu2024CVPR,zhang2025training}), which can be categorized as pixel-level classification models. These methods assign a binary label to each pixel~(either forged or authentic) based on discriminative features, and generate a corresponding localization mask. In contrast, we take a generative perspective, aiming to synthesize the most likely forgery localization map conditioned on the input image and its manipulation traces. To this end, we propose a novel framework that, for the first time, integrates the generative power of SD into a forensic setting, enabling effective and accurate image forgery localization.

In this paper, we first provide a theoretical rationale for applying image generation models~(such as SD) to the task of image forgery localization. We argue that compressing an image into the latent space inevitably leads to the loss of certain visual details, particularly high-frequency signals that are often indicative of forging artifacts. By treating these high-frequency signals as an additional modality and introducing them into the latent space, the probability of accurately generating forgery localization results can be increased from a probabilistic perspective. Training the model to maximize this probability enables effective image forgery localization. Building upon this theoretical foundation, we introduce a generalizable paradigm that applies generative image models to the task of image forgery localization.

Subsequently, we leverage the multi-modal architecture of Stable Diffusion V3 (SD3)~\footnote{\url{https://stability.ai/news/introducing-stable-diffusion-3-5}}. In this architecture, text and image modalities are fused within a shared latent space, enabling cross-modal generation. Typically, the text modality serves as a conditioning signal to guide the content or style of the generated image. Inspired by this mechanism, we propose treating image forgery residuals, namely high-frequency features extracted using specifically designed high-pass filters, as an explicit additional modality to replace the text modality. These residual signals, which are highly sensitive to forging artifacts, are injected into the latent space as conditioning signals during the training process. In this way, our model thereby produces an image forgery localization map.

Extensive experiments demonstrate that our proposed method consistently achieves state-of-the-art (SOTA) performance across multiple benchmark datasets. Furthermore, we validate the model's generalization ability using real-world forgery images, including natural scene forgeries and document image forgeries, and assess its robustness through various post-processing techniques and social media processing. Encouragingly, the model exhibits strong generalization and robustness, maintaining superior performance even on unseen real-world forgery datasets, which highlights the effectiveness and practical applicability of our approach. Our main contributions are threefold:

\begin{itemize}
\item We present the first theoretical demonstration that integrating supplementary information into generative image models enables effective image forgery localization. Furthermore, we propose a generalizable paradigm for image forensic models by adapting latent variable space-based generative architectures to forensic tasks.
\item We introduce a pioneering approach to pixel-level forgery localization using Stable Diffusion (SD) by incorporating forensic residuals as a latent-space modality. This method leverages SD's semantic knowledge while effectively capturing subtle manipulation traces.
\item Experimental results demonstrate that our approach achieves state-of-the-art (SOTA) performance and robustness across multiple benchmark natural scene datasets, with superior generalization to real-world images.
\end{itemize}

The rest of the paper is organized as follows. Sect.~\ref{sec:proposed} describes our proposed method, Sect.~\ref{sec:exp} details experiments and results, Sect.~\ref{sec:ablation} covers ablation experiments and results, and Sect.~\ref{sec:conclude} provides a conclusion.

\section{Proposed Method}
\label{sec:proposed}
\label{Proposed Method}

In this section, we first present our motivation and theoretically demonstrate the feasibility of applying SD to image forensics. Based on the demonstration, we show how to construct the loss function. We also propose a general paradigm for adapting latent variable space-based generative models into forensic models. We then provide an overview of the proposed framework and offer detailed descriptions of each component.

\subsection{Preliminaries}
\label{Preliminaries}

SD is a robust latent diffusion model that generates high-quality images by learning from latent representations, bypassing direct pixel-space operations. Trained on extensive image-text datasets, SD employs a pre-trained variational autoencoder (VAE)~\cite{pinheiro2021variational} to encode images into latent features. Its diffusion process iteratively introduces noise into the latent space and learns to denoise it, guided by a text encoder conditioned on textual prompts.

SD assumes that each image sample \(\boldsymbol{X}\) is drawn i.i.d. from an unknown generative process involving latent variables \(\boldsymbol{Z}\). The generation process is typically modeled in two steps:
\begin{enumerate}
    \item Sample latent variables: \(\boldsymbol{Z} \sim p_{\theta^*}(\boldsymbol{Z})\);
    \item Generate observations via an iterative denoising process: \(\boldsymbol{X} \sim p_{\theta^*}(\boldsymbol{X} \mid \boldsymbol{Z})\).
\end{enumerate}

Here, the prior \(p_{\theta^*}(\boldsymbol{Z})\) and the conditional likelihood \(p_{\theta^*}(\boldsymbol{X} \mid \boldsymbol{Z})\) are assumed to belong to parameterized families \(p_{\theta}(\boldsymbol{Z})\) and \(p_{\theta}(\boldsymbol{X} \mid \boldsymbol{Z})\), respectively, which are differentiable with respect to both \(\theta\) and \(\boldsymbol{Z}\). In practice, however, the true parameters \(\theta^*\) and the latent variables \(\boldsymbol{Z}\) are unobserved. Researchers typically optimize \(\theta\) by maximizing the Evidence Lower Bound (ELBO)~\cite{Diederik2014_auto}, and approximate each \(\boldsymbol{Z}\) using an encoder.

\subsection{Motivation}
\label{Motivation}

To enable the application of SD to image forgery localization, we adapt the theoretical model such that the generated output \(\boldsymbol{X}_{output} = \left(d_{i,j,k}\right)^{(w \times h \times c)},\quad d_{i,j,k} \in [0, 255]\) no longer represents an image, but instead corresponds to a forgery localization result \(\boldsymbol{M} = \left(m_{i,j}\right)^{(w \times h)},\quad m_{i,j} \in \{0, 1\}\), where \(w\), \(h\), and \(c\) denote the width, height, and number of channels of the image, respectively. In \(\boldsymbol{M}\), a pixel value of 0 indicates that the pixel is authentic, while a value of 1 indicates that the pixel is forged. The main challenge, then, is how to increase the probability of generating accurate forgery localization results.

When using encoders, such as those in VAE frameworks, to compress the original image $\boldsymbol{X}$ into a latent representation \(\boldsymbol{Z}\), the image undergoes a substantial compression ratio. For instance, a \(512 \times 512 \times 3\) image may be reduced to a \(64 \times 64 \times 16\) latent representation.

This process typically involves a series of convolutional layers with stride values greater than 1, which downsample the spatial dimensions. Simultaneously, the number of output channels is incrementally increased across layers, transforming the original image $\boldsymbol{X} = (d_{i,j,k})^{w \times h \times c}$ into a latent representation $\boldsymbol{Z} = (z_{i,j,k})^{w' \times h' \times c'}$. Here, $w'$, $h'$, and $c'$ denote the downsampled width, height, and channel dimensions, respectively, where $w' = w/s$, $h' = h/s$, and $c' \gg c$ in most architectures, with $s$ being the downsampling factor.

In the frequency domain, this operation causes spectral folding, namely periodic replication of the frequency content, as given by:

\begin{equation}
\scalebox{0.85}{$
\mathcal{F}\{\boldsymbol{Z}\}(\omega_1, \omega_2) = \frac{1}{s^2} \sum_{k_1=0}^{s-1} \sum_{k_2=0}^{s-1} \mathcal{F}\{\boldsymbol{X}\} \left( \frac{\omega_1 + 2\pi k_1}{s}, \frac{\omega_2 + 2\pi k_2}{s} \right)
$}
\end{equation}

where \(\mathcal{F}\{\cdot\}\) denotes the 2D discrete Fourier transform, \((\omega_1, \omega_2)\) are the spatial frequency variables, and \(k_1, k_2 \in \{0, \dots, s-1\}\) are the folding indices. After downsampling, the frequency range is compressed from \([-\pi, \pi]\) to \([-\pi/s, \pi/s]\).

According to the Nyquist--Shannon Sampling Theorem~\footnote{\href{https://en.wikipedia.org/wiki/Nyquist\%E2\%80\%93Shannon_sampling_theorem}{https://en.wikipedia.org/wiki/Nyquist-Shannon\_sampling\_theorem}}, \emph{a continuous signal that is bandlimited to a maximum spatial frequency can be perfectly reconstructed from its discrete samples if and only if the sampling rate $s$ satisfies $\pi/s$ being greater than the aforementioned maximum spatial frequency}.

This means that if either $\omega_1$ or $\omega_2$ exceeds $\pi/s$, the corresponding high-frequency information will be irretrievably lost due to aliasing during downsampling.

In the context of visual representation, this implies that fine-grained details such as object textures and edges may be lost, whereas global semantic features and structural layouts are more likely to be preserved \cite{higgins2017beta, zhao2019infovae}.

However, in image forgery tasks, forged regions often exhibit subtle yet discriminative artifacts, such as local texture inconsistencies, high-frequency structural discontinuities, and edge anomalies. These artifacts are particularly prominent along the boundaries between manipulated objects and the background.

To compensate for the high-frequency information that is lost in \(\boldsymbol{Z}\), we introduce an auxiliary feature \(\boldsymbol{F}\): high-frequency residual signals, where forgery traces are typically more concentrated. These provide valuable cues for forgery localization. Next, we provide an information-theoretic justification that incorporating high-frequency information \(\boldsymbol{F}\) increases the probability of accurately predicting the forgery localization map \(\boldsymbol{M}\).

The forgery localization target \(\boldsymbol{M}\), containing only two values, 0 and 1. This indicates that the pixel values in \(\boldsymbol{M}\) are discontinuous along the region boundaries, which, in the mathematical sense, implies the presence of high-frequency components. Therefore, forgery localization can be interpreted as detecting or recovering high-frequency structures that correspond to semantic edges introduced by manipulation artifacts.

Assuming the model aims to learn the conditional distribution \( p(\boldsymbol{M} \mid \boldsymbol{Z}) \), our objective is to enhance prediction quality. In this context, the mutual information \( I(\boldsymbol{Z}; \boldsymbol{M}) \) quantifies how much information about \( \boldsymbol{M} \) is retained in \( \boldsymbol{Z} \). A higher \( I(\boldsymbol{Z}; \boldsymbol{M}) \) indicates that the latent representation contains more relevant information for accurately localizing manipulations.

The mutual information before and after introducing the additional information \( \boldsymbol{F} \) can be formulated as \( I(\boldsymbol{Z}; \boldsymbol{M}) \) and \( I(\boldsymbol{Z,F}; \boldsymbol{M}) \), which can be decomposed as follows:

\begin{equation}
I(\boldsymbol{Z};\boldsymbol{M}) = H(\boldsymbol{M}) - H(\boldsymbol{M}|\boldsymbol{Z})
\label{eq:mutual1}
\end{equation}

\begin{equation}
I(\boldsymbol{Z},\boldsymbol{F};\boldsymbol{M}) = H(\boldsymbol{M}) - H(\boldsymbol{M}|\boldsymbol{Z},\boldsymbol{F})
\label{eq:mutual2}
\end{equation}

Where \(H(\boldsymbol{M})\) is the entropy of \(\boldsymbol{M}\) (which remains constant), so the difference depends on the conditional entropy. Since the forgery localization mask \(\boldsymbol{M}\) inherently contains high-frequency components. By introducing high-frequency residual information \(\boldsymbol{F}\), the model receives complementary edge-aware cues. From an information-theoretic perspective, these residuals reduce the conditional entropy:

\begin{equation}
H(\boldsymbol{M}|\boldsymbol{Z},\boldsymbol{F}) \leq H(\boldsymbol{M}|\boldsymbol{Z})
\label{eq:mutual3}
\end{equation}

By substituting Equations \eqref{eq:mutual3} into Equations \eqref{eq:mutual1} and \eqref{eq:mutual2}, we obtain:

\begin{equation}
I(\boldsymbol{Z},\boldsymbol{F};\boldsymbol{M}) > I(\boldsymbol{Z};\boldsymbol{M})
\label{eq:mutual4}
\end{equation}

Therefore, incorporating \( \boldsymbol{F} \) enhances the informativeness of the latent representation with respect to the target output \( \boldsymbol{M} \), leading to improved performance in forgery localization.

\subsection{Loss Construction}
\label{Loss Construction}

The aforementioned theoretical demonstration offers a solution for adapting image generation models into image forgery localization models. We now turn to the design of an effective loss function that can guide the model to produce accurate and high-fidelity forgery localization results. Specifically, our objective is to maximize the log-likelihood of the observed localization mask \(\boldsymbol{M}\) given \(\boldsymbol{X}\) and \(\boldsymbol{F}\):

\begin{equation}
\scalebox{0.85}{$
\log p(\boldsymbol{M} \mid \boldsymbol{X}, \boldsymbol{F}) = \log \int p(\boldsymbol{M} \mid \boldsymbol{Z, X, F})\, p(\boldsymbol{Z} \mid \boldsymbol{X, F})\, d\boldsymbol{Z}
$}
\end{equation}

Using the decomposition of the joint probability, $p(\boldsymbol{M, Z \mid X, F}) = p(\boldsymbol{M \mid Z, X, F}) p(\boldsymbol{Z \mid X, F})$:

\begin{equation}
\label{eq:original}
\scalebox{0.85}{$
\log p(\boldsymbol{M} \mid \boldsymbol{X}, \boldsymbol{F}) = \log \int p(\boldsymbol{M, Z \mid X, F})\, d\boldsymbol{Z}
$}
\end{equation}

Since computing the integral in Equation \eqref{eq:original} is intractable, we can introduce a trainable variational distribution $q(\boldsymbol{Z \mid X, F, M})$ to approximate the posterior. Starting from the following identity:
\begin{equation}
\scalebox{0.85}{$
\log p(\boldsymbol{M \mid X, F}) = \log \int q(\boldsymbol{Z \mid X, F, M}) \frac{p(\boldsymbol{M, Z \mid X, F})}{q(\boldsymbol{Z \mid X, F, M})} d\boldsymbol{Z}
$}
\end{equation}

According to the definition of expectation, this integral is equivalent to taking the expectation of the function $\frac{p(\boldsymbol{M, Z \mid X, F})}{q(\boldsymbol{Z \mid X, F, M})}$ with respect to the distribution $q(\boldsymbol{Z \mid X, F, M})$:
\begin{equation}
\scalebox{0.8}{$
\int q(\boldsymbol{Z \mid X, F, M}) \frac{p(\boldsymbol{M, Z \mid X, F})}{q(\boldsymbol{Z \mid X, F, M})} d\boldsymbol{Z} = \mathbb{E}_{q(\boldsymbol{Z \mid X, F, M})} \left[ \frac{p(\boldsymbol{M, Z \mid X, F})}{q(\boldsymbol{Z \mid X, F, M})} \right]
$}
\end{equation}
Therefore, we obtain:

\begin{equation}
\scalebox{0.85}{$
\log p(\boldsymbol{M \mid X, F}) = \log \mathbb{E}_{q(\boldsymbol{Z \mid X, F, M})} \left[ \frac{p(\boldsymbol{M, Z \mid X, F})}{q(\boldsymbol{Z \mid X, F, M})} \right]
$}
\end{equation}

Applying Jensen's inequality ($\log \mathbb{E}[\cdot] \geq \mathbb{E}[\log(\cdot)]$), we get:
\begin{equation}
\label{eq:lowbound}
\scalebox{0.85}{$
\log p(\boldsymbol{M \mid X, F}) \geq \mathbb{E}_{q(\boldsymbol{Z \mid X, F, M})} \left[ \log \frac{p(\boldsymbol{M, Z \mid X, F})}{q(\boldsymbol{Z \mid X, F, M})} \right]
$}
\end{equation}

Observing Equation \eqref{eq:lowbound}, we can conclude that the probability of predicting \(\boldsymbol{M}\) given \(\boldsymbol{X}\) and \(\boldsymbol{F}\) has a lower bound, which is exactly the ELBO. ELBO provides a variational approximation to the true likelihood by introducing a tractable posterior, leading to a decomposition into a reconstruction term and a regularization term. This allows the model to learn latent variables \(\boldsymbol{Z}\) that capture forgery-relevant features while ensuring distributional consistency via Kullback–Leibler (KL) divergence. Again using the decomposition of the joint probability, $p(\boldsymbol{M, Z \mid X, F}) = p(\boldsymbol{M \mid Z, X, F}) p(\boldsymbol{Z \mid X, F})$:
\begin{equation}
\scalebox{0.85}{$
\text{ELBO} = \mathbb{E}_{q(\boldsymbol{Z \mid X, F, M})} \left[ \log \frac{p(\boldsymbol{M \mid Z, X, F})\, p(\boldsymbol{Z \mid X, F})}{q(\boldsymbol{Z \mid X, F, M})} \right]
$}
\end{equation}

We now split the logarithm inside the expectation:
\begin{equation}
\label{eq:ELBO}
\scalebox{0.85}{$
\begin{split}
\text{ELBO} &= \mathbb{E}_{q} \left[ \log p(\boldsymbol{M} \mid \boldsymbol{Z}, \boldsymbol{X}, \boldsymbol{F}) \right] \\
&+ \mathbb{E}_{q} \left[ \log p(\boldsymbol{Z} \mid \boldsymbol{X}, \boldsymbol{F}) - \log q(\boldsymbol{Z} \mid \boldsymbol{X}, \boldsymbol{F}, \boldsymbol{M}) \right]
\end{split}
$}
\end{equation}

Note that the second expectation term is exactly the negative KL divergence. So, Equation \eqref{eq:ELBO} can be written as:

\begin{equation}
\label{eq:ELBO_final}
\scalebox{0.85}{$
\begin{split}
\text{ELBO} = \mathbb{E}_{q(\boldsymbol{Z \mid X, F, M})} [\log p(\boldsymbol{M \mid Z, X, F})] \\
 - D_{\mathrm{KL}}(q(\boldsymbol{Z \mid X, F, M}) \parallel p(\boldsymbol{Z \mid X, F}))
\end{split}
$}
\end{equation}

By negating the ELBO, we obtain $L$. When $L$ decreases, the ELBO increases, consequently leading to a higher probability of predicting \(\boldsymbol{M}\) given \(\boldsymbol{X}\) and \(\boldsymbol{F}\). The expectation on the right-hand side of Equation \eqref{eq:ELBO_final} represents the model's ability to reconstruct the observed variable \(\boldsymbol{M}\), i.e., the reconstruction loss. Thus, the model training objective becomes minimizing the sum of the reconstruction loss and the KL divergence:
\begin{equation}
\scalebox{0.85}{$
\begin{split}
L = -\mathbb{E}_{q(\boldsymbol{Z \mid X, F, M})} [\log p(\boldsymbol{M \mid Z, X, F})] \\
+ D_{\mathrm{KL}}(q(\boldsymbol{Z \mid X, F, M}) \parallel p(\boldsymbol{Z \mid X, F}))
\end{split}
$}
\end{equation}

While this theoretical objective provides a probabilistic foundation, its direct implementation requires specific design. In the actual implementation, we employ two loss functions with reference to \cite{huo2025generative}: a latent space matching loss \(\mathcal{L}_{lm} \) and an image-space localization loss \(\mathcal{L}_{loc} \). \(\mathcal{L}_{loc} \) corresponds to the reconstruction loss in the above objective $L$, while \(\mathcal{L}_{lm} \) corresponds to the KL divergence. \(\mathcal{L}_{lm} \) encourages similarity between the latent representation \(\boldsymbol{Z}_m \) derived from the real mask and the predicted latent representation \(\hat{\boldsymbol{Z}}_m \). It is formulated as follows:

\begin{equation}
\mathcal{L}_{lm} =  | \boldsymbol{Z}_m - \hat{\boldsymbol{Z}}_m |_2^2
\end{equation}

\(\mathcal{L}_{loc} \) ensures that the predicted mask \(\hat{\boldsymbol{M}} \) aligns closely with the real mask \(\boldsymbol{M} \), providing direct supervision in the image space. This loss is defined using the soft Dice loss formulation:

\begin{equation}
\mathcal{L}_{loc} = 1 - \frac{2 \cdot \sum (\boldsymbol{M} \circ \hat{\boldsymbol{M}})}{\sum \boldsymbol{M} + \sum \hat{\boldsymbol{M}}}
\end{equation}

where \( \circ \) denotes element-wise multiplication. The overall loss function used during training is the sum of these two components:

\begin{equation}
\mathcal{L} = \mathcal{L}_{lm} + \mathcal{L}_{loc}
\end{equation}

The following sections introduce an easy-to-implement yet effective instantiation of the proposed framework, specifically detailing the implementation of the model.

\subsection{The Overall Framework}
\label{Localization}

\begin{figure*}[!t]
  \centering
  \begin{overpic}[width=0.6\textwidth]{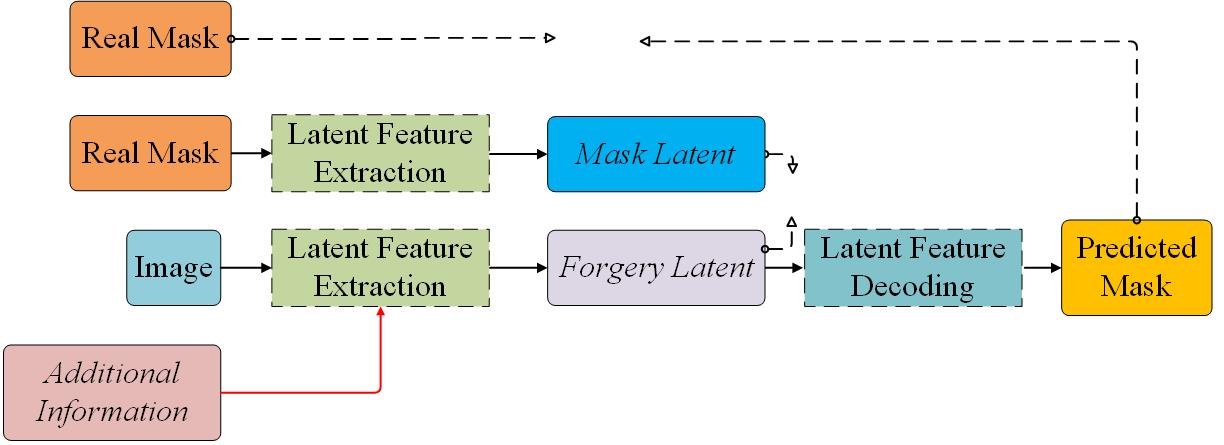}
    \put(46,33){\small $\mathcal{L}_{loc}$} 
    \put(63,20){\small $\mathcal{L}_{lm}$} 
  \end{overpic}
  \caption{General paradigm for applying image generation models to image forgery localization.}
  \label{fig_-1}
\end{figure*}

In this section, we propose a unified framework, as shown in Figure \ref{fig_-1}.

The light green boxes indicate two identical latent variable extractors. The one in the middle extracts the latent representation from the real mask, resulting in Mask Latent, while the one at the bottom, together with the light red additional information injector, generates the augmented latent representation of the image with additional information, referred to as Forgery Latent. The previously mentioned loss term \(\mathcal{L}_{lm}\) is computed based on Mask Latent and Forgery Latent.

Forgery Latent is passed through a latent decoder to obtain the predicted mask, which, together with the real mask, is used to compute the aforementioned loss term \(\mathcal{L}_{loc}\).

In theory, the latent feature extraction and decoding in the figure can adopt the encoder and decoder from any existing image generation model. The additional information can be derived from existing forgery feature extraction modules.

Based on this paradigm, we propose an image forgery localization model built upon SD3. In the following two subsections, we introduce how to utilize the encoder and decoder of SD3 to compress an image into the latent space and decode it to obtain the predicted mask. We then describe how to incorporate high-frequency residual information—strongly correlated with forgery traces—as additional information into the latent space.

\subsection{Image Forgery Localization Model Based on SD3}
\label{Localization}

\begin{figure*}[!t]
  \centering
  \begin{overpic}[width=1\textwidth]{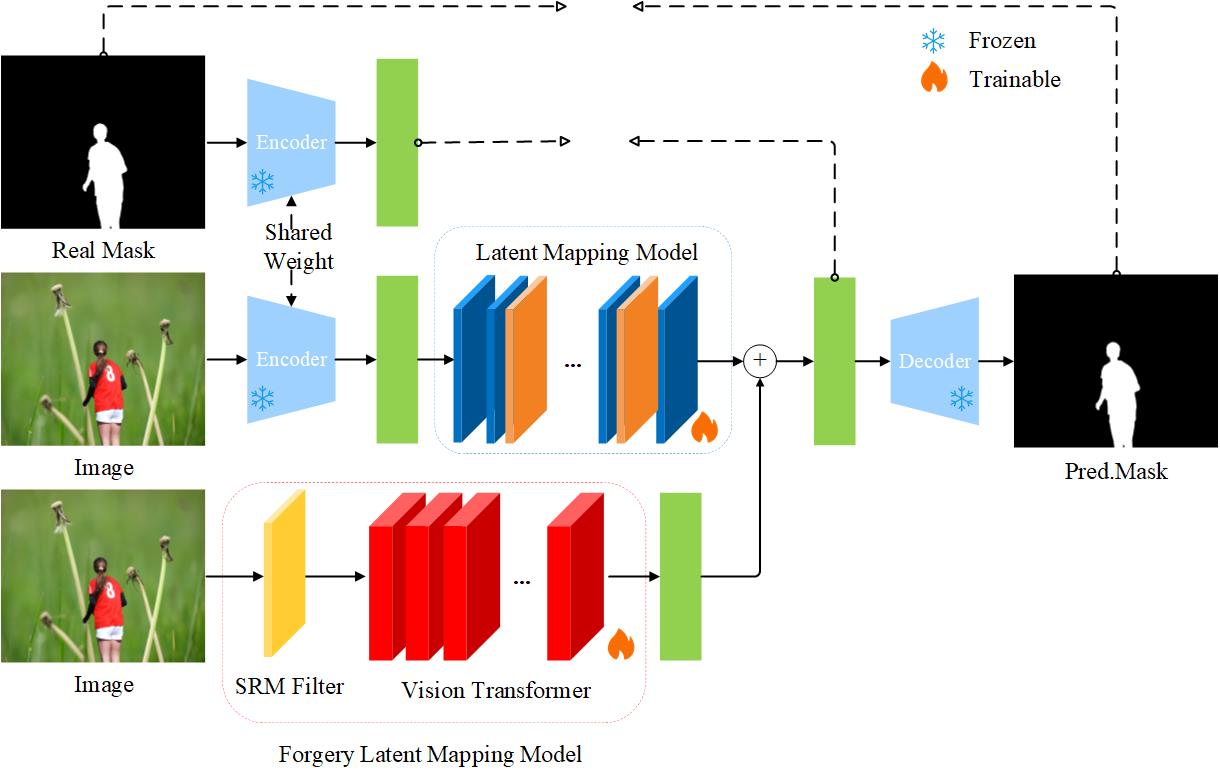}
    \put(47,63){\large $\mathcal{L}_{loc}$}  
    \put(47,52){\large $\mathcal{L}_{lm}$} 
    \put(31,44){\small $\boldsymbol{Z}_m$} 
    \put(31,26){\small $\boldsymbol{Z}_i$} 
    \put(54,8){\small $\boldsymbol{Z}_f$} 
    \put(67,25){\small $\hat{\boldsymbol{Z}}_m$} 
    
  \end{overpic}
  \caption{The concrete implementation of our proposed framework for forgery image localization.}
  \label{fig_-2}
\end{figure*}
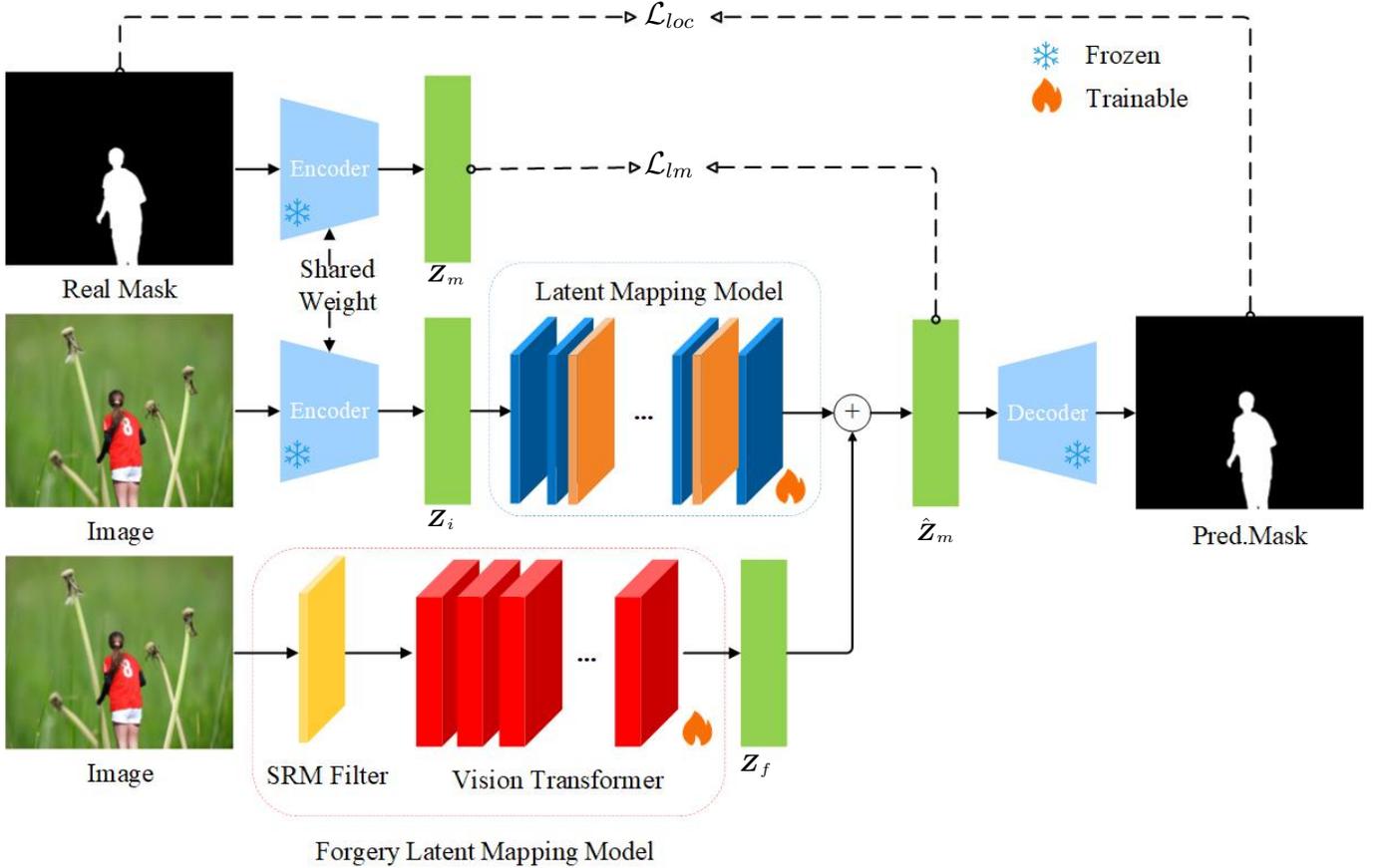

\begin{figure}[!t]
    \centering
    \subfloat[\footnotesize Detailed structure of the Latent Mapping Model]{%
        \includegraphics[width=0.9\linewidth]{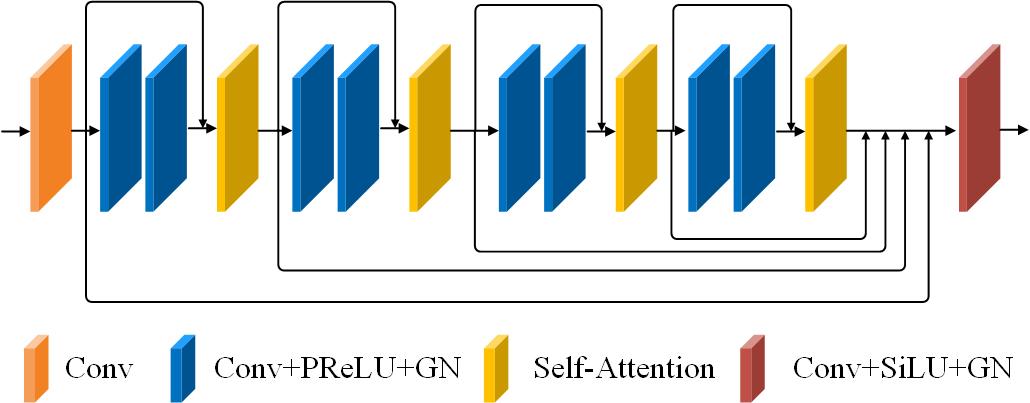}%
        \label{fig:subfig-a}
    }\hfill

    \subfloat[\footnotesize Detailed structure of the Forgery Latent Mapping Model]{%
        \includegraphics[width=0.9\linewidth]{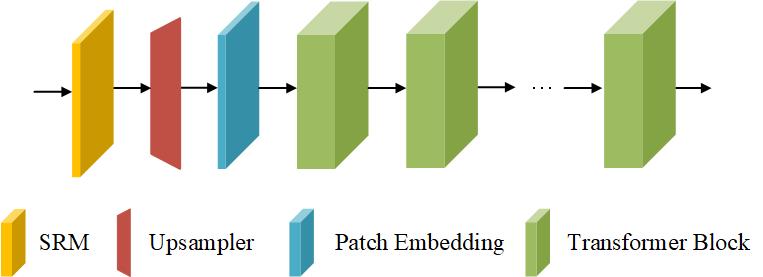}%
        \label{fig:subfig-b}
    }

    \caption{The concrete implementation of Latent Mapping Model and Forgery Latent Mapping Model.}
    \label{fig_-3}
\end{figure}

The proposed model's structure is shown in Figure \ref{fig_-2}. 

In our model, the pre-trained encoder and decoder from SD3 are utilized as an image tokenizer, transforming images into latent representations and vice versa. Selecting a well-matched pair of 
encoder and decoder is essential to ensure that the latent space accurately preserves both image content and mask semantics, which is critical for downstream tasks.

We follow the approach in \cite{huo2025generative} and adopt the VAE module from SD3, which has been trained on large-scale natural image datasets \cite{schuhmann2022laion}, endowing it with rich latent representations and strong zero-shot generalization ability. The SD3-VAE achieves near-lossless image reconstruction \cite{Rombach_2022_CVPR}, making it a strong candidate for image forgery localization tasks.

Given a forged image and its corresponding forgery localization mask, a frozen VAE encoder is used to extract the latent representations \(\boldsymbol{Z}_i \) and \(\boldsymbol{Z}_m \). Before incorporating the additional information, we process the latent variables of the forged image using \textit{Latent Mapping Model} (LMM) \cite{huo2025generative} (as shown in the blue box in the Figure \ref{fig_-2}), which aims to preserve the contours of the objects in the image, as these contours are strongly correlated with the forgery localization results. We designed the \textit{Forgery Latent Mapping Model} (FLMM) (as shown in the red box in the Figure \ref{fig_-2}) to obtain the latent representation of the forgery traces, denoted as \(\boldsymbol{Z}_f \). The feature \(\boldsymbol{Z}_f \) serves as additional information and is fused with \( \boldsymbol{Z}_i \). In this work, we adopt matrix concatenation to obtain the fused latent representation \(\hat{\boldsymbol{Z}}_m \). Finally, \(\hat{\boldsymbol{Z}}_m \) is decoded by the pre-trained decoder to produce the forgery localization result in the image space. Note that both the encoder and decoder are pre-trained and kept frozen during training and inference.

\subsection{Forgery Latent Mapping Model}
\label{Forgery Latent Mapping Model}

The additional information plays a crucial role in enhancing the model's ability to accurately predict forgery localization results. Given that manipulation traces are often hidden within the high-frequency components of an image, we apply Spatially Rich Model (SRM) \cite{fridrich2012rich} filters to effectively extract these subtle and imperceptible clues. SRM filters are a set of steganalysis filters designed to detect statistical anomalies introduced by data hiding (such as noise distribution anomalies and local correlation disruptions), and can therefore also be used to identify digital forging traces (e.g., edge discontinuities, noise pattern disturbances, and local texture inconsistencies) \cite{Zhou_2018_CVPR}. To match the dimensions of the latent variables, we use three filters, as shown in Equation \eqref{eq:-1}. 

\begin{equation}
\scalebox{0.8}{%
  $\begin{array}{c} 
    \begin{aligned}
      & \frac{1}{4}
      \begin{bmatrix}
      0 & 0 & 0 & 0 & 0 \\
      0 & -1 & 2 & -1 & 0 \\
      0 & 2 & -4 & 2 & 0 \\
      0 & -1 & 2 & -1 & 0 \\
      0 & 0 & 0 & 0 & 0
      \end{bmatrix}
      \quad
      \frac{1}{12}
      \begin{bmatrix}
      -1 & 2 & -2 & 2 & -1 \\
      2 & -6 & 8 & -6 & 2 \\
      -2 & 8 & -12 & 8 & -2 \\
      2 & -6 & 8 & -6 & 2 \\
      -1 & 2 & -2 & 2 & -1
      \end{bmatrix}
    \end{aligned} \\[2ex] 
    \begin{aligned}
      & \frac{1}{2}
      \begin{bmatrix}
      0 & 0 & 0 & 0 & 0 \\
      0 & 0 & 0 & 0 & 0 \\
      0 & 1 & -2 & 1 & 0 \\
      0 & 0 & 0 & 0 & 0 \\
      0 & 0 & 0 & 0 & 0
      \end{bmatrix}
    \end{aligned}
  \end{array}$}
\label{eq:-1}
\end{equation}

To seamlessly incorporate the extracted residual information into the latent space, we adopt the Vision Transformer \cite{dosovitskiy2021an} for feature extraction.

Vision Transformer is commonly employed for tasks such as image classification, object detection, and semantic segmentation, and has been adopted as the encoder in the Segment Anything Model (SAM) \cite{kirillov2023segment} to enhance global feature extraction capabilities. We leverage the pre-trained encoder weights from SAM to enhance the representational capacity of our model \footnotemark[1]. To adapt the large model to our specific task of image forgery localization, we fine-tune the encoder of the SAM so that it can better capture manipulation traces and extract more discriminative features related to forged regions.

\footnotetext[1]{https://github.com/facebookresearch/segment-anything}

This design enables the latent representation to retain critical manipulation traces necessary for accurate localization. The detailed structures of the Latent Manipulation Module (LMM) and the Forgery Latent Mapping Model (FLMM) are illustrated in Figure~\ref{fig_-3}. By introducing SRM-extracted high-frequency residuals as an auxiliary modality in this manner, we significantly enhance the model's capacity to perceive, retain, and localize manipulation artifacts that might otherwise be lost during latent encoding.

\section{Experiments}
\label{sec:exp}

\subsection{Experimental Setup}
\label{ch:Datasets}

\begin{table}[!t]
\caption{Dataset Composition.}
\label{tab_Dataset}
\centering
\begin{tabular}{|c||c||c|}
\hline
Dataset&Images&Operation\\
\hline
    tampCOCO \cite{kwon2021cat} & 822776& Copy-move, Splice\\
    IMD2020 \cite{novozamsky2020imd2020} & 2009& Copy-move, Splice, Inpaint\\
    CASIA1 \cite{dong2013casia} & 920&  Copy-move, Splice\\
    CASIA2 \cite{dong2013casia} & 5123&  Copy-move, Splice\\
    NIST16 \cite{guan2019mfc} & 564&  Copy-move, Splice, Inpaint\\
    Coverage \cite{wen2016coverage} & 100&  Copy-move\\
    Columbia \cite{ng2004data} & 160 & Splice\\
    MISD \cite{MISD2021} & 302 & Splice\\
    OSNs \cite{wu2022robust} & 1684& Post-processing on social media\\
    CocoGlide \cite{guillaro2023trufor} & 512&  Generate\\
    GRE \cite{sun2024GRE} & 492& Generate\\
    ACDTamp & 337& Copy-move, Splice, Inpaint\\
    PS-boundary \cite{zhuang2021image} & 1000& Real-world PS-based manipulations\\
    PS-arbitrary \cite{zhuang2021image} & 1000& Unknown PS-based manipulations\\
\hline
\end{tabular}
\end{table}

The datasets involved in the experiments of this paper are shown in Table \ref{tab_Dataset}. To maintain consistent experimental conditions in the comparative experiments, we used the same training set as in \cite{guillaro2023trufor}: tampCoco \cite{kwon2021cat}, IMD2020 \cite{novozamsky2020imd2020}, and CASIA2 \cite{dong2013casia} datasets. In total, these datasets contain approximately 830,000 manipulated and pristine images with varied forgery patterns. The image resolutions typically range from 384\(\times\)256 to 1920\(\times\)1080. These datasets were selected due to their diversity in manipulation types (such as splicing, copy-move, and inpainting), the realism of the editing artifacts, and their widespread use in existing image forgery localization benchmarks.

In subsequent experiments, we consistently used the weights obtained from this training set for a series of tests. The test sets included NIST16 \cite{guan2019mfc}, Coverage \cite{wen2016coverage}, CASIA1 \cite{dong2013casia}, Columbia \cite{ng2004data}, MISD \cite{MISD2021}, OSNs \cite{wu2022robust}, CocoGlide \cite{guillaro2023trufor}, GRE \cite{sun2024GRE}, ACDTamp, PS-boundary \cite{zhuang2021image}, and PS-arbitrary \cite{zhuang2021image} datasets.

NIST16, Coverage, Columbia, MISD, and CASIA1 are widely used in the literature and contain a variety of image manipulation, including copy-move, splice, and inpainting operations. These datasets have been widely utilized to assess the performance of image forgery localization techniques. The OSNs dataset, comprises forged images originating from Columbia, NIST16, DSO \cite{de2013exposing}, and CASIA1. These images have been processed by transmission through various social media platforms, including Facebook, Weibo, WeChat, and WhatsApp. The CocoGlide, and GRE datasets, however, are specifically designed to include forged regions generated by the diffusion model, which is known for producing highly realistic forged images. 

\begin{figure}[!t]
\centering
\includegraphics[width=0.5\textwidth]{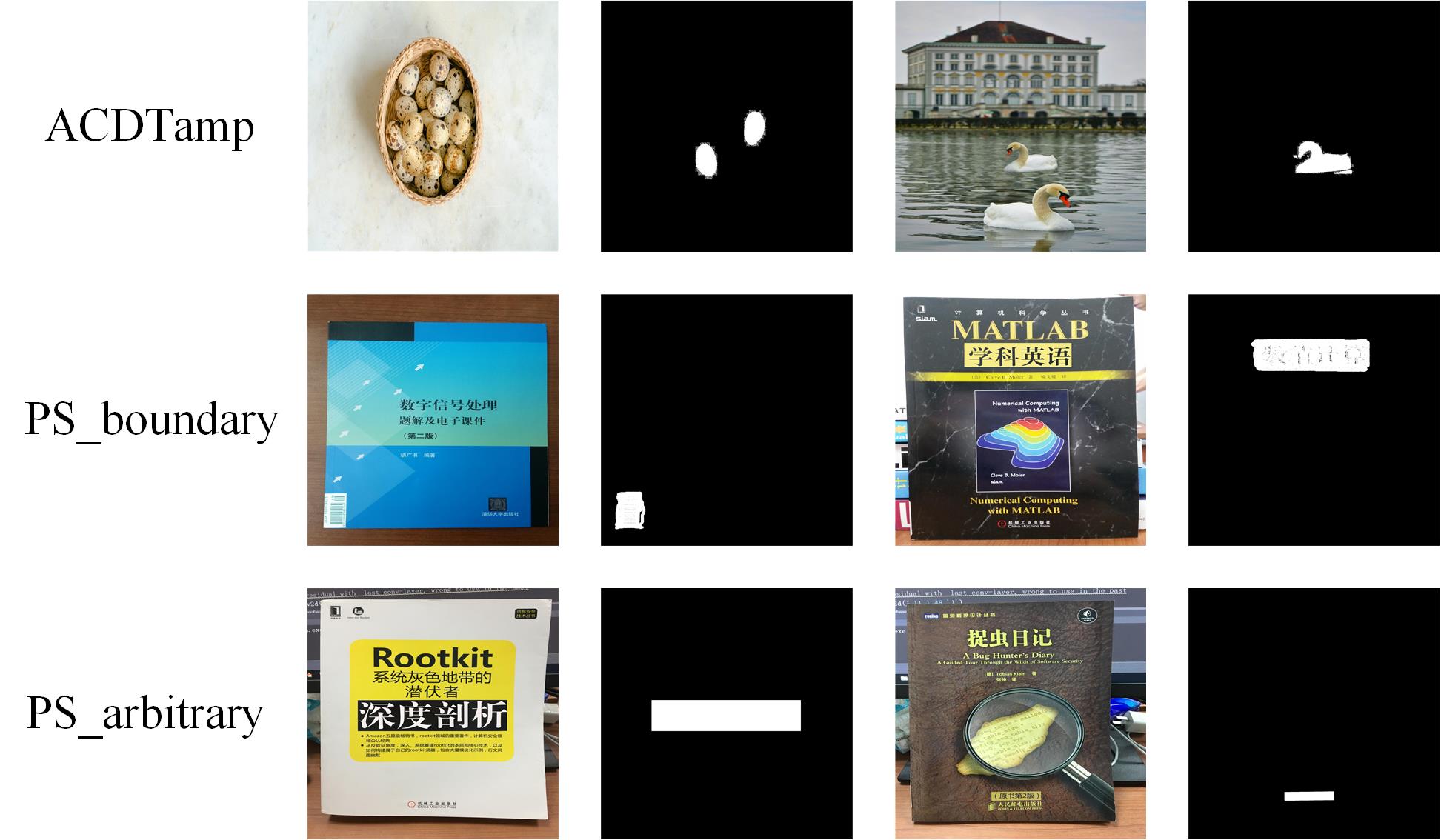}
\caption{Example images from our lab's proprietary real forgery dataset.}
\label{fig_-4}
\end{figure}

The ACDTamp, PS-boundary \cite{zhuang2021image}, and PS-arbitrary \cite{zhuang2021image} datasets are constructed by our laboratory to simulate real-world forgery scenarios, comprising a total of 2,337 images. The ACDTamp dataset consists of natural scene images collected from the web, which were subsequently manipulated using the ACDSee software. The PS-boundary and PS-arbitrary datasets contain forged book cover images created by different individuals using Photoshop, involving boundary-constrained and arbitrary-shaped forging, respectively. Sample images from these datasets are shown in Figure~\ref{fig_-4}.

It is noteworthy that there is no overlap between the training and testing datasets, ensuring the integrity and fairness of the evaluation process.

The output of image forgery localization is a binary mask, where forged pixels are labeled as 1 (positive) and authentic pixels as 0 (negative). Each pixel prediction falls into one of four categories: True Positive (TP), True Negative (TN), False Positive (FP), or False Negative (FN). To evaluate model performance, we adopt the Intersection over Union (IoU) and F1 score. IoU measures the overlap between the predicted and ground truth tampered regions, while F1 balances both the FP and FN rates via Precision and Recall:

\begin{equation}
\text{Precision} = \frac{\text{TP}}{\text{TP} + \text{FP}}
\end{equation}

\begin{equation}
\text{Recall} = \frac{\text{TP}}{\text{TP} + \text{FN}}.
\end{equation}

The formulas for IoU and F1 score are given in Equations~\eqref{eq:4} and \eqref{eq:5}:

\begin{equation}
\text{IoU} = \frac{\text{TP}}{\text{TP} + \text{FP} + \text{FN}}
\label{eq:4}
\end{equation}

\begin{equation}
F_1 = 2 \cdot \frac{\text{Precision} \cdot \text{Recall}}{\text{Precision} + \text{Recall}}
\label{eq:5}
\end{equation}

In the task of forgery localization, the distinction between forged and original regions is often ambiguous. If the final result effectively differentiates forged and original regions - even if their labels are reversed - it remains valid. Following the approach in \cite{guillaro2023trufor}, we compute two F1 scores: one for the localization result and another for its complement. The higher of these two values is adopted as the final F1 score.

We used the VAE backbone from SD3\footnotemark[2], and the training was conducted on 2 NVIDIA A100-80G GPUs with a batch size of 4. The AdamW optimizer was employed with a learning rate of 1e-4 and a warm-up period of 5 epochs. F1 score was computed using a fixed threshold of 0.5.

\footnotetext[2]{https://huggingface.co/stabilityai/stable-diffusion-3-medium}

\subsection{Experimental results on traditional forgeries}
\label{ch:Cross-database}

\begin{figure*}[!t]
\centering
\includegraphics[width=1\textwidth]{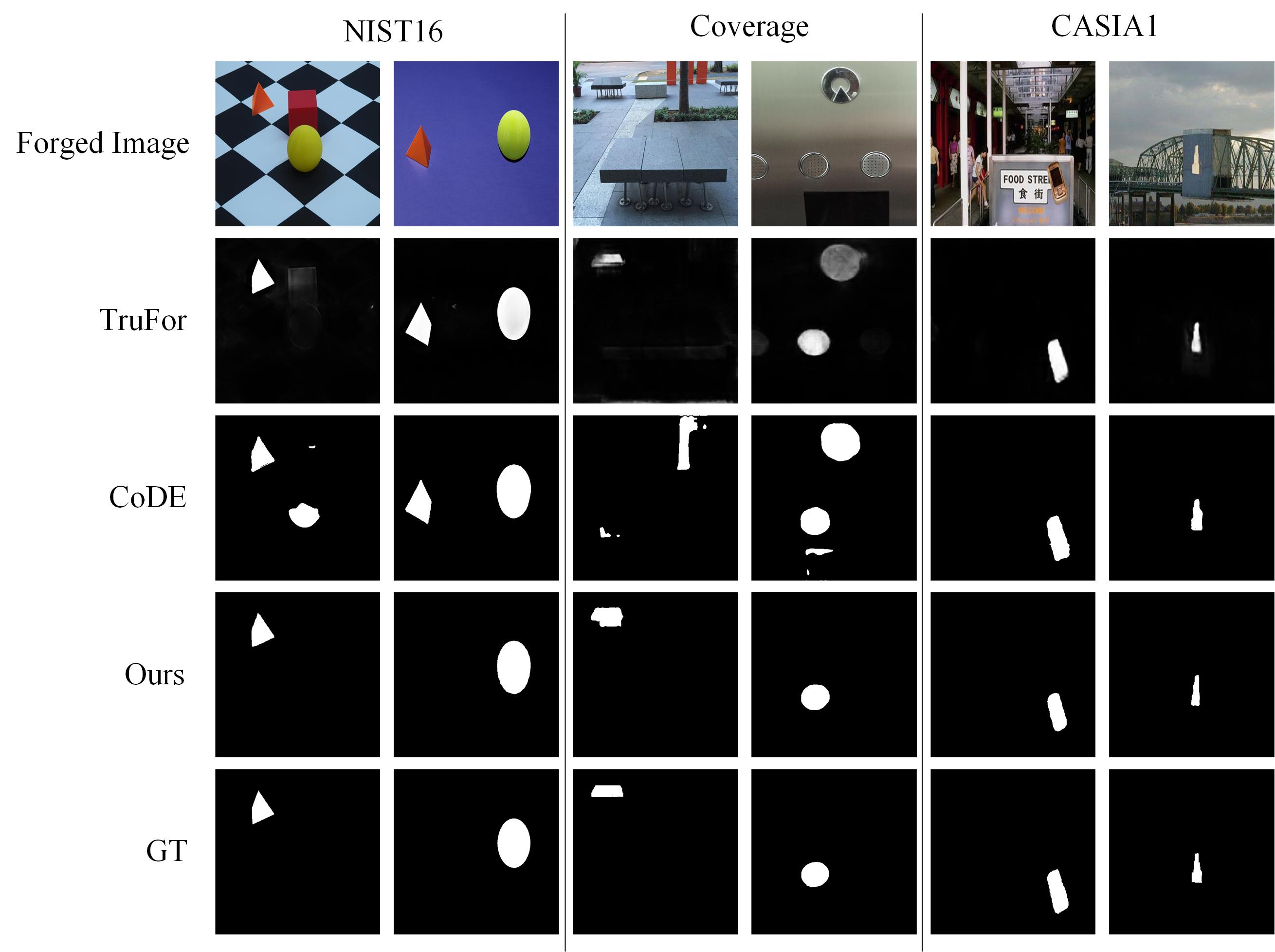}
\caption{Example of traditional forgery localization visual results.}
\label{fig_-5}
\end{figure*}

In this section, we evaluate on three datasets generated by traditional forgery methods: NIST16, Coverage, and CASIA1. We selected 7 algorithms for comparison, including the latest deep learning-based approaches: CAT-Net v2 \cite{kwon2021cat}, IF-OSN \cite{wu2022robust}, MVSS \cite{chen2021image}, PSCC-Net \cite{liu2022pscc}, TruFor \cite{guillaro2023trufor}, CoDE \cite{Peng2024CoDE}, and SAM-LORA. SAM-LORA is the Low-Rank Adaptation (LORA)-finetuned SAM framework introduced in our comparative experiments, designed to validate the effectiveness of the standalone LORA adaptation strategy for downstream image forgery localization tasks. The SAM-LORA method involves adding a small auxiliary network to each transformer layer in the encoder of SAM \cite{hu2022lora}. We freeze both the encoder and decoder of SAM and only train the added auxiliary network. SAM uses the ViT-B model\footnotemark[1]. Comparative results are presented in Table \ref{tab_Cross}, with representative forgery localization visualizations shown in Figure \ref{fig_-5}. For demonstration, we selected two sample images from each of the NIST16, Coverage, and CASIA1 datasets.

\begin{table*}[!t]
\caption{IoU and F1 score performance of traditional image forgery localization(The results are computed using a fixed threshold 0.5).}
\label{tab_Cross}
\centering
\begin{tabular}{|c||c|c||c|c||c|c||c|c||c|c|}
\hline
\multirow{2}{*}{Method} & \multicolumn{2}{c||}{CASIA1} & \multicolumn{2}{c||}{Coverage} & \multicolumn{2}{c||}{NIST16} & \multicolumn{2}{c||}{Columbia} & \multicolumn{2}{c|}{MISD}\\
\cline{2-11}
 & IoU & F1 & IoU & F1 & IoU & F1 & IoU & F1 & IoU & F1\\
\hline
    CAT-Net & 0.622 & 0.703 & 0.232 & 0.290 & 0.230 & 0.301 & 0.742 & 0.792 & 0.562 & 0.683\\
    IF-OSN & 0.465 & 0.509 & 0.180 & 0.266 & 0.252 & 0.331 & 0.614 & 0.713  & 0.447 & 0.568\\
    MVSS-Net & 0.379 & 0.423 & 0.381 & 0.458 & 0.248 & 0.305 & 0.588 & 0.677  & 0.484 & 0.627\\
    PSCC-Net & 0.538 & 0.627 & 0.282 & 0.394 & 0.227 & 0.298 & 0.825 & \underline{0.866} & 0.475 & 0.605\\
    TruFor & 0.629 & 0.693 & 0.451 & \underline{0.525} & 0.291 & 0.362 & 0.748 & 0.807 & 0.425 & 0.643\\
    CoDE & \underline{0.637} & \underline{0.723} & 0.362 & 0.464 & 0.339 & 0.420 & 0.844 & 0.881 & \underline{0.615} & \underline{0.738}\\
    SAM-LORA & 0.625 & 0.684 & 0.457 & 0.510 & \underline{0.426} & \underline{0.481} & \underline{0.838} & 0.860 & 0.563 & 0.689\\
    Ours & \textbf{0.673} & \textbf{0.732} & \textbf{0.476} & \textbf{0.586} & \textbf{0.436} & \textbf{0.503} & \textbf{0.851} & \textbf{0.901} & \textbf{0.619} & \textbf{0.743}\\
\hline
\end{tabular}
\end{table*}

Overall, our proposed method achieves the best performance across all datasets, leading in both IoU and F1 score in most cases, demonstrating its strong generalization ability in forgery localization tasks.

When compared with CAT-Net, IF-OSN, and MVSS-Net, our proposed method exhibits a clear advantage on the Coverage, NIST16, and Columbia datasets. On the Coverage dataset, our method achieves an F1 score of 0.586, outperforming CAT-Net (0.290), IF-OSN (0.266), and MVSS-Net (0.458) by up to 29.6 percentage points. On NIST16, our method attains an F1 score of 0.503, which is substantially higher than CAT-Net (0.301), IF-OSN (0.331), and MVSS-Net (0.305) by at least 17.2 percentage points. Similarly, on the Columbia dataset, our method outperforms the best PSCC-Net with an IoU of 0.851 and an F1 of 0.901, compared to PSCC-Net’s IoU of 0.825 and F1 of 0.866, showcasing a notable performance gap.

In comparison to recent state-of-the-art methods like TruFor, CoDE, and SAM-LORA, our proposed method consistently outperforms the competition across most datasets. Specifically, it surpasses TruFor by 6.1 points in F1 on the Coverage dataset (0.586 vs. 0.525) and by 14.1 points in F1 on NIST16 (0.503 vs. 0.362). On Columbia, our method exceeds TruFor by 0.103 in IoU (0.851 vs. 0.748) and 0.094 in F1 (0.901 vs. 0.807). Additionally, our method leads on the MISD dataset with an F1 score of 0.743, compared to CoDE’s 0.738. This consistent superiority across all datasets underscores the robust performance of our approach.

As can be seen from Figure \ref{fig_-5}, the localization results of TruFor and CoDE on the NIST16 and Coverage datasets appear sparse and demonstrate significant false positives and missed regions, failing to precisely identify forged areas. In contrast, our method produces more precise localization masks, identifying the forgery areas more accurately with fewer false positives. The results highlight the superior performance of our method in localizing image forgeries compared to other methods, as demonstrated by the clearer and more accurate detection of manipulated regions, which closely aligns with the ground truth.

\subsection{Experimental results on diffusion-based forgeries}
\label{ch:diffusion}

\begin{figure}[!t]
\centering
\includegraphics[width=0.5\textwidth]{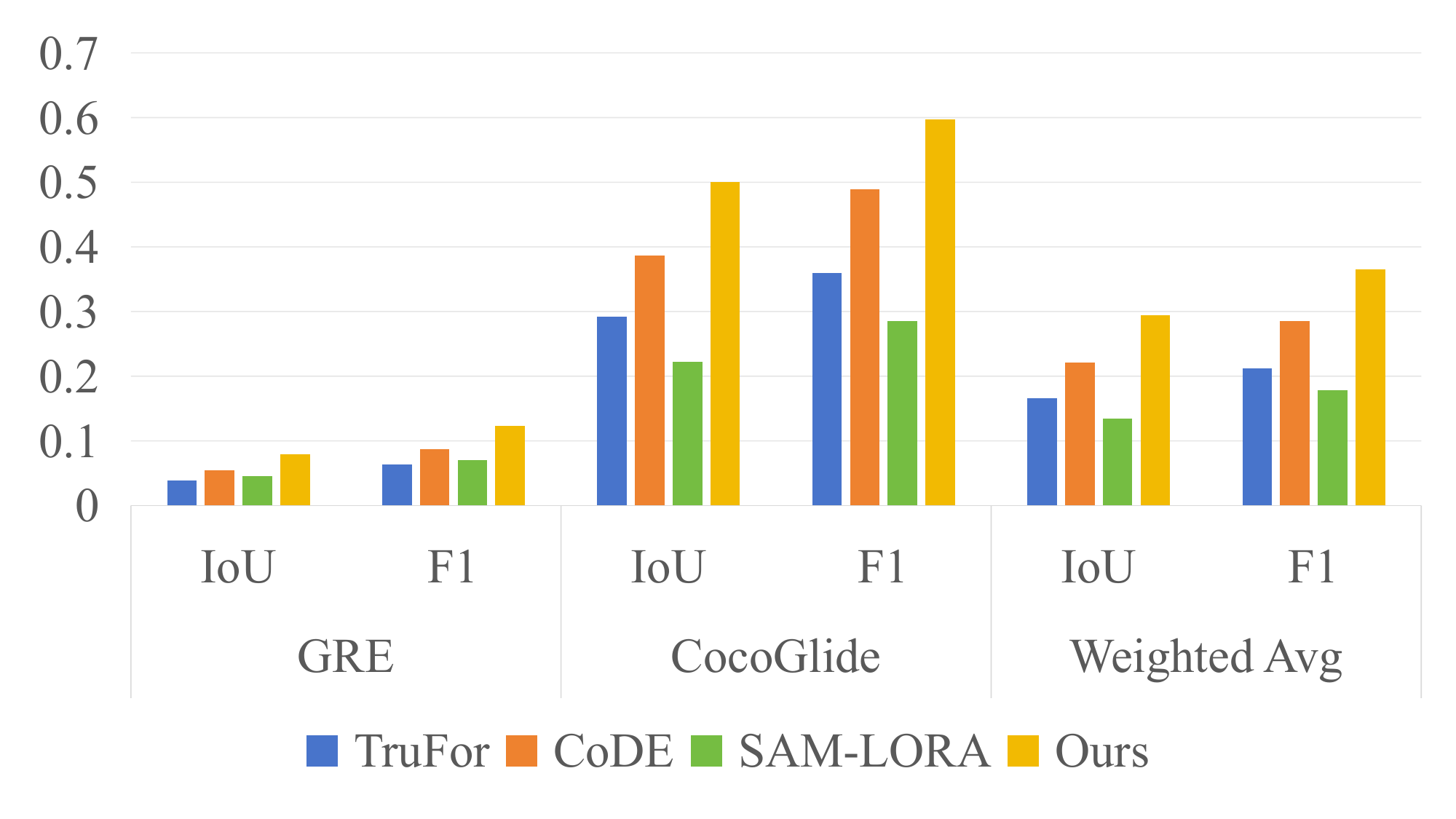}
\caption{IoU and F1 performance of diffusion-based forgery localization.}
\label{fig_-6}
\end{figure}

In this section, we evaluate on two datasets generated using diffusion models: CocoGlide and GRE. Three algorithms are selected for comparison: TruFor, CoDE, and SAM-LORA. The experimental results are presented in Figure \ref{fig_-6}, which demonstrates that our method achieves unprecedented performance on diffusion-generated datasets, surpassing all baseline methods by substantial margins. Specifically, it obtains the highest weighted average IoU (0.294) and F1-score (0.365), significantly outperforming prior approaches such as CoDE (0.221/0.285) and TruFor (0.166/0.212). On the challenging GRE dataset, where all methods exhibit relatively low scores, our approach still achieves the best IoU (0.079) and F1-score (0.123), indicating superior robustness under complex tampering scenarios.

Furthermore, on the CocoGlide dataset, which contains high-quality diffusion-based forgeries, our method demonstrates substantial improvements, reaching 0.5 IoU and 0.597 F1-score, surpassing all baselines by a notable margin. These results confirm that our approach generalizes well to emerging diffusion forgeries.

\subsection{Generalization Analysis}
\label{ch:Generalization}

The evaluation of generalization capability remains crucial in image forgery detection. To rigorously assess our method's adaptability to practical scenarios, we conduct comprehensive tests across three real-world forgery datasets: ACDTamp, PS-boundary, and PS-arbitrary datasets. While ACDTamp contains natural scene images consistent with the training distribution, both PS-boundary and PS-arbitrary exclusively comprise book cover images – a domain absent from the training data.

\begin{figure}[!t]
\centering
\includegraphics[width=0.5\textwidth]{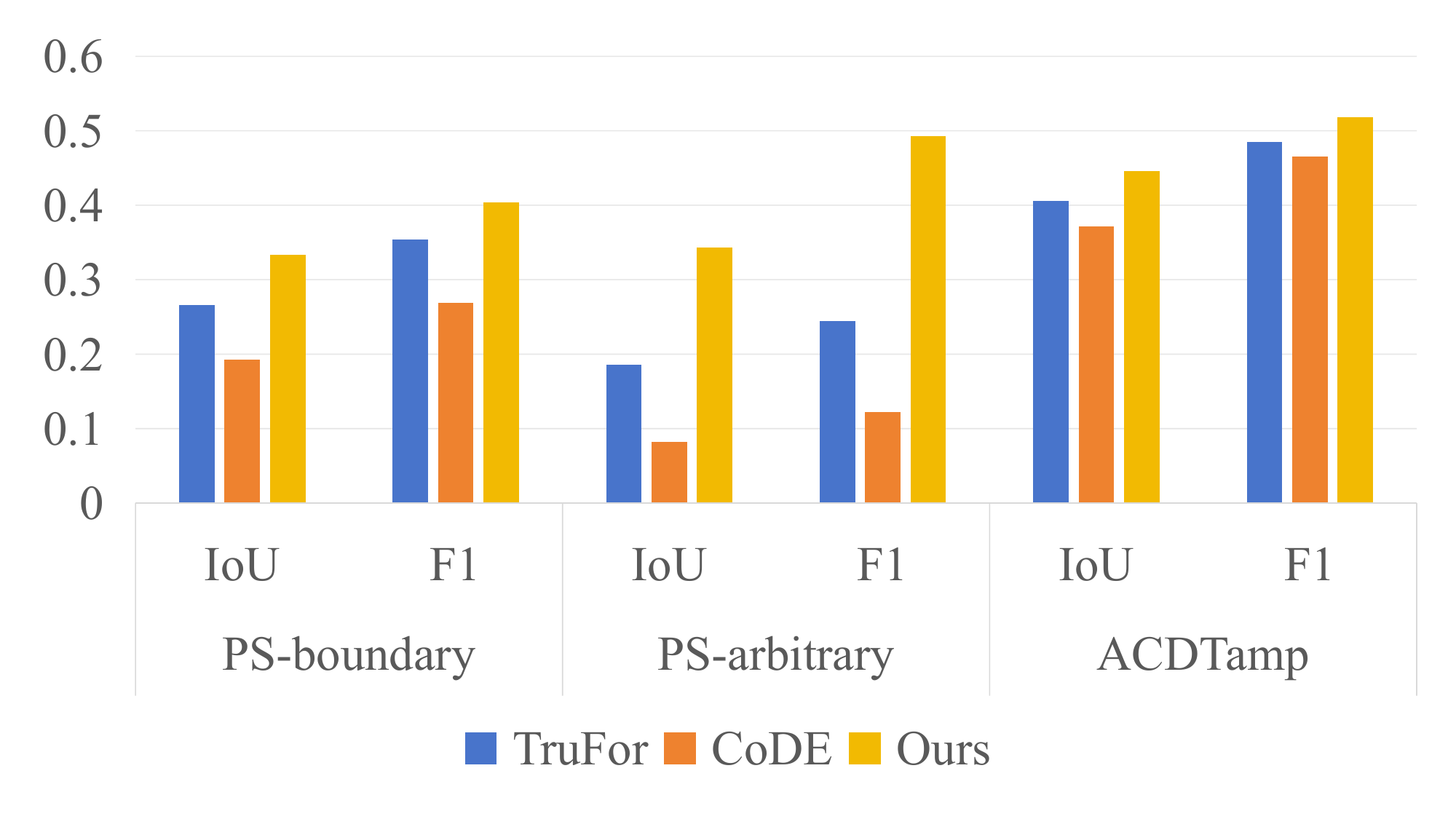}
\caption{IoU and F1 performance of generalization performance experiment.}
\label{fig_-7}
\end{figure}

As presented in Figure \ref{fig_-7}, our method consistently outperforms prior approaches across all three subsets of the dataset. On the PS-boundary subset, which features tampering along semantic object boundaries, our approach achieves the highest IoU (0.333) and F1-score (0.404), outperforming TruFor (0.266/0.354) and CoDE (0.193/0.269), indicating more accurate localization of structured manipulations.

For the more challenging PS-arbitrary subset, where forged regions exhibit irregular and free-form boundaries, our method demonstrates a substantial advantage with an IoU of 0.343 and F1-score of 0.493. In contrast, the baseline methods perform poorly under these conditions, with CoDE scoring as low as 0.082 (IoU) and 0.122 (F1), highlighting the robustness of our approach to less constrained manipulations.

Finally, on the ACDTamp dataset, which includes more realistic scenarios, our method again achieves the best results (IoU: 0.446, F1: 0.518), significantly outperforming TruFor (0.406/0.485) and CoDE (0.372/0.465). This demonstrates the strong generalization ability of our model when applied to complex, real-world image forgeries.

\subsection{Robustness Against Post-processing Operations}
\label{ch:Post-processing}

We selected three types of post-processing operations: adding Gaussian noise, JPEG compression, and resizing. For each operation, three different parameter levels were defined. Specifically, the Gaussian noise had a mean of 0 and standard deviations of 0.1, 0.3, and 0.5. JPEG compression quality levels were set to 70, 80, and 90. Scaling factors were set to 0.7, 0.8, and 0.9.

Gaussian noise with a mean of 0 ensures that the noise is symmetrically distributed around zero, a common assumption in many image processing. This type of noise does not introduce systematic bias to pixel values. A standard deviation of 0.1 represents mild noise, suitable for simulating slight image degradation. A standard deviation of 0.3 indicates moderate noise, resembling real-world conditions, especially for images captured in low-light environments. A standard deviation of 0.5 represents significant noise, simulating extreme cases of severe image degradation. Similarly, the three levels of JPEG compression correspond to varying image quality from low to high after compression. The three scaling factors were designed to better adapt to common resizing operations required for different display sizes or platform requirements.

\begin{figure*}[!t]
\centering
\includegraphics[width=1\textwidth]{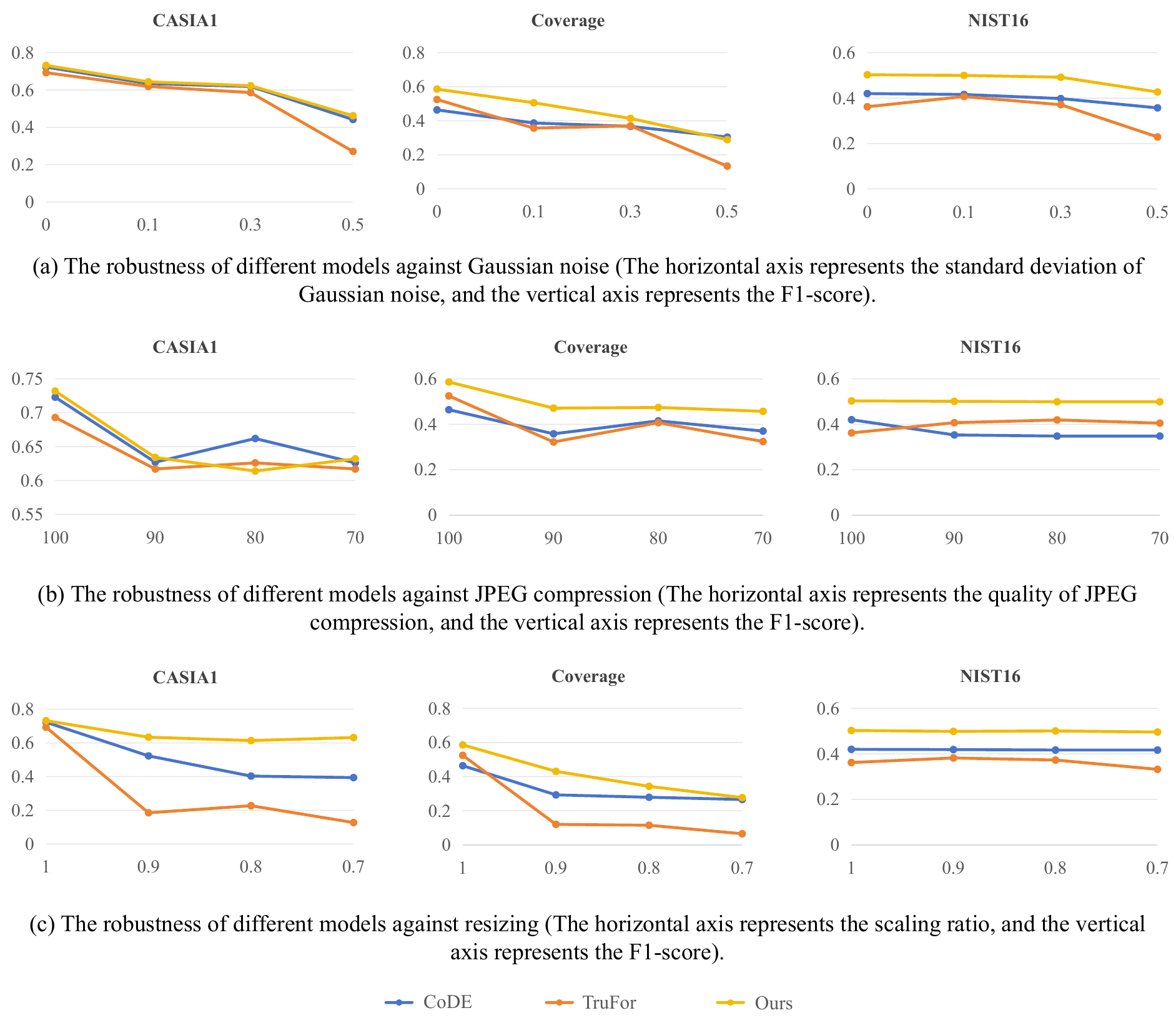}
\caption{The robustness of different models.}
\label{fig_-8}
\end{figure*}

We used three datasets for validation: CASIA1, NIST16, and Coverage datasets. The robustness performance of the models across these datasets is illustrated in Figures \ref{fig_-8}.

As shown in subfigures (a)–(c), our method consistently achieves higher F1-scores than CoDE and TruFor under increasing levels of Gaussian noise, JPEG compression, and resizing distortions, demonstrating greater resilience to common image degradation scenarios.

Notably, under severe perturbation conditions—such as low JPEG quality (e.g., 70), high noise levels, or significant downscaling (scale ratio = 0.7)—the performance of baseline methods degrades sharply, while our method maintains relatively stable F1-scores across all datasets. This indicates that our approach generalizes better under practical, low-quality image conditions and is thus more suitable for real-world forensic applications where post-processing is prevalent.

\subsection{Robustness Analysis on Online Social Networks}
\label{ch:Social Networks}

Online Social Networks (OSNs) introduce complex and unpredictable conditions for image forgery localization. During the upload and sharing process, images are typically subjected to multiple degradations such as compression, resizing, and platform-specific post-processing. These operations not only reduce image quality but also obscure or distort subtle tampering artifacts, making it significantly more difficult to detect manipulation traces accurately. As a result, the robustness of traditional detection methods is greatly challenged in such real-world settings.

As shown in Table \ref{tab_Social}, performance degradation is observed across all methods when images are compressed by online social networks (OSNs), but our proposed method maintains exceptional robustness. Specifically, under various OSN conditions such as Facebook, Weibo, and WeChat, our method outperforms all baselines, achieving the highest IoU and F1 scores on datasets like Columbia, NIST16, and CASIA1. For instance, on Facebook, it achieves an F1 score of 0.499 on NIST16, and on WeChat, it achieves 0.482 on the same dataset, showcasing superior performance under compression.

When images are compressed through WhatsApp, known for its heavy compression, our method continues to show its strength. It achieves the highest F1 scores across multiple datasets, including 0.896 on Columbia, 0.496 on NIST16, and 0.402 on DSO, outpacing other methods. In contrast, methods like CAT-Net and TruFor show inconsistent performance across datasets, with CAT-Net excelling only on Columbia but falling short on others, indicating its limited generalization under compression.

Overall, the robust performance of our method across all OSNs and datasets demonstrates its strong generalization ability. Unlike other models, which struggle with aggressive compression, our method consistently delivers superior results, highlighting its practical applicability for real-world scenarios where media is often subjected to complex transmission and compression transformations.

\begin{table*}[!t]
\caption{IoU and F1 performance of robustness analysis on online social networks(The results are computed using a fixed threshold 0.5).}
\label{tab_Social}
\centering
\begin{tabular}{|c||c||c|c||c|c||c|c||c|c|}
\hline
\multirow{2}{*}{Method}& \multirow{2}{*}{OSN} & \multicolumn{2}{c||}{Columbia} & \multicolumn{2}{c||}{NIST16} & \multicolumn{2}{c||}{DSO} & \multicolumn{2}{c|}{CASIA1}\\
\cline{3-10}
 & & IoU & F1 & IoU & F1 & IoU & F1 & IoU & F1\\
\hline
    MVSS-Net & None & 0.589         & 0.677             & 0.248             & 0.305             & 0.184             & 0.262             & 0.379             & 0.432  \\
    PSCC-Net & None & 0.825         & 0.866             & 0.227             & 0.298             & 0.136             & 0.200             & 0.538             & 0.627 \\
    CAT-Net & None & 0.742          & 0.792             & 0.230             & 0.301             & 0.402             & 0.478             & 0.622             & 0.703 \\
    IF-OSN & None & 0.614           & 0.713             & 0.252             & 0.331             & 0.317             & 0.445             & 0.465             & 0.509  \\
    TruFor & None & 0.748           & 0.807             & 0.291             & 0.362             & \textbf{0.865}    & \textbf{0.910}    & 0.629             & 0.693 \\
    CoDE & None & 0.844             & 0.881             & 0.339             & 0.420             & 0.265             & 0.380             & 0.637             & 0.723 \\
    Ours & None & \textbf{0.851}    & \textbf{0.901}    & \textbf{0.436}    & \textbf{0.503}    & 0.314             & 0.402             & \textbf{0.673}    & \textbf{0.732} \\
\hline
    MVSS-Net & Facebook & 0.604        & 0.692          & 0.179          & 0.226          & 0.186          & 0.263          & 0.316          & 0.368  \\
    PSCC-Net & Facebook & 0.792        & 0.838          & 0.216          & 0.286          & 0.130          & 0.192          & 0.499          & 0.584 \\
    CAT-Net & Facebook & \textbf{0.891}& \textbf{0.916} & 0.119          & 0.151          & 0.098          & 0.121          & 0.522          & 0.627 \\
    IF-OSN & Facebook & 0.614          & 0.716          & 0.250          & 0.326          & 0.330          & 0.458          & 0.419          & 0.464  \\
    TruFor & Facebook & 0.671          & 0.749          & 0.262          & 0.333          & \textbf{0.571} & \textbf{0.674} & 0.606          & 0.673 \\
    CoDE & Facebook & 0.845            & 0.882          & 0.331          & 0.411          & 0.276          & 0.392          & 0.609          & 0.699 \\
    Ours & Facebook & 0.875            & 0.895          & \textbf{0.439} & \textbf{0.499} & 0.331          & 0.420          & \textbf{0.613} & \textbf{0.702} \\
\hline
    MVSS-Net & Weibo & 0.588        & 0.678          & 0.151          & 0.192          & 0.173          & 0.242          & 0.341          & 0.389  \\
    PSCC-Net & Weibo & 0.358        & 0.406          & 0.114          & 0.160          & 0.061          & 0.092          & 0.427          & 0.506 \\
    CAT-Net & Weibo & \textbf{0.888}& \textbf{0.920} & 0.157          & 0.208          & 0.015          & 0.023          & 0.341          & 0.422 \\
    IF-OSN & Weibo & 0.628          & 0.726          & 0.215          & 0.290          & 0.256          & 0.375          & 0.421          & 0.466  \\
    TruFor & Weibo & 0.731          & 0.800          & 0.251          & 0.321          & \textbf{0.375} & \textbf{0.478} & 0.576          & 0.638 \\
    CoDE & Weibo & 0.845            & 0.883          & 0.325          & 0.408          & 0.268          & 0.382          & \textbf{0.613} & \textbf{0.701} \\
    Ours & Weibo & 0.867            & 0.880          & \textbf{0.431} & \textbf{0.489} & 0.314          & 0.401          & 0.606          & 0.680 \\
\hline
    MVSS-Net & Wechat & 0.597        & 0.684          & 0.135          & 0.172          & 0.48           & 0.205          & 0.207          & 0.245  \\
    PSCC-Net & Wechat & 0.789        & 0.834          & 0.199          & 0.268          & 0.127          & 0.189          & 0.405          & 0.490 \\
    CAT-Net & Wechat & 0.802         & 0.847          & 0.148          & 0.191          & 0.011          & 0.017          & 0.105          & 0.139 \\
    IF-OSN & Wechat & 0.638          & 0.732          & 0.209          & 0.281          & 0.251          & 0.367          & 0.358          & 0.405  \\
    TruFor & Wechat & 0.702          & 0.773          & 0.260          & 0.342          & \textbf{0.323} & \textbf{0.446} & 0.508          & 0.570 \\
    CoDE & Wechat & 0.840           & 0.878          & 0.322          & 0.413          & 0.257          & 0.368          & 0.536           & 0.629 \\
    Ours & Wechat & \textbf{0.875}   & \textbf{0.899} & \textbf{0.422} & \textbf{0.482} & 0.320         & 0.408         & \textbf{0.585}   & \textbf{0.664} \\
\hline
    MVSS-Net & WhatsApp & 0.597        & 0.687          & 0.107          & 0.138          & 0.122          & 0.171          & 0.291          & 0.338  \\
    PSCC-Net & WhatsApp & 0.791        & 0.842          & 0.224          & 0.295          & 0.124          & 0.183          & 0.482          & 0.574 \\
    CAT-Net & WhatsApp & 0.891         & 0.920          & 0.168          & 0.201          & 0.017          & 0.022          & 0.356          & 0.420 \\
    IF-OSN & WhatsApp & 0.628          & 0.727          & 0.239          & 0.313          & 0.233          & 0.340          & 0.431          & 0.478  \\
    TruFor & WhatsApp & 0.667          & 0.747          & 0.301          & 0.383          & 0.299          & 0.389          & 0.599          & 0.664 \\
    CoDE & WhatsApp & 0.844            & 0.883          & 0.342          & 0.420          & 0.265          & 0.377          & 0.606         & 0.697 \\
    Ours & WhatsApp & \textbf{0.875}   & \textbf{0.896} & \textbf{0.433} & \textbf{0.496} & \textbf{0.313} & \textbf{0.402} & \textbf{0.627} & \textbf{0.705} \\
\hline
\end{tabular}
\end{table*}

%
%
%

\section{Ablation Study}
\label{sec:ablation}

To evaluate the individual impact of each design choice in our method, we conducted systematic ablation studies.

First, we removed both the SRM and the FLMM module, which effectively means no additional information was introduced; the model relied solely on the latent variables of the forged image to generate the forgery localization results.

Next, we removed both the VAE and LMM modules, meaning the model no longer utilized the latent variables of the forged image and instead relied solely on the high-frequency components of the input image to generate localization results.

Then, we did not load the pre-trained weights for the LMM module, which effectively means the LMM had to be trained from scratch without leveraging prior knowledge. This helps assess the importance of pre-trained boundary information in the final performance.

Finally, we retained the VAE but removed the LMM module. This means the model did not extract boundary-related information from the latent space and instead utilized all latent features of the forged image to perform localization. We trained the model using the IMD2020 dataset and tested it using the NIST16 dataset.

Table \ref{ablation} presents the results of a series of ablation studies designed to assess the contribution of each component in our proposed framework. When both the SRM and FLMM modules were removed, the F1-score dropped to 0.342, demonstrating the importance of incorporating high-frequency priors and frequency-aware feature separation. When both the VAE and LMM modules were removed, resulting in the model relying solely on handcrafted high-frequency features, performance further declined to 0.356, indicating that handcrafted priors alone are insufficient for accurate localization.

Furthermore, excluding the LMM module but retaining the VAE led to a moderate improvement (F1 = 0.389), suggesting that although latent representations are useful, the lack of boundary-aware information limits their effectiveness. Training the LMM module from scratch without pre-trained weights also resulted in performance degradation (F1 = 0.363), underscoring the benefit of transferring boundary priors from external data. In contrast, the complete model achieved the best performance with an F1-score of 0.421, validating the complementary strengths of each component and the overall design of our architecture.

\begin{table}[!t]
\caption{The results of the ablation experiments.}
\label{ablation}
\centering
\begin{tabular}{|c||c|}
\hline
Method&F1\\
\hline
    w/o SRM+FLMM                 & 0.342\\
    w/o VAE+LMM            & 0.356\\
    w/o LMM                    & 0.389\\
    w/o LMM Pre-trained        & 0.363\\
    Complete Model              & \textbf{0.421}\\
\hline
\end{tabular}
\end{table}

\section{Concluding remarks}
\label{sec:conclude}

This study is the first to integrate the generative capabilities of Stable Diffusion into image forensics for efficient forgery localization. The main contributions are: (1) a theoretical demonstration of the feasibility of applying SD’s generative power to forensic tasks; (2) the use of the SD3 multi-modal framework, where high-frequency image residuals are treated as an explicit modality and introduced into the latent space as a conditioning signal to enhance localization performance.

Experimental results demonstrate that our method achieves SOTA performance across multiple public datasets, effectively detecting both traditional forging operations and those conducted using generative models. Additionally, our model exhibits strong robustness and generalization ability, making it more adaptable to real-world scenarios.

For future work, we aim to focus on two key areas: 1. Developing computationally efficient variants through targeted knowledge distillation, with the goal of maintaining forgery localization accuracy while reducing the model's operational costs; 2. Exploring the integration of the noise-adding and denoising processes from Stable Diffusion to further improve the model’s performance. These improvements are anticipated to simplify the overall pipeline and significantly enhance the model’s capability to localize forgeries, particularly in challenging real-world environments.

%
\bibliography{paper.bib}
\bibliographystyle{IEEEtran}

\vfill

\end{document}